\documentclass{article}
\pdfoutput=1

\PassOptionsToPackage{numbers, compress}{natbib}
\usepackage[preprint]{neurips_2023}

\usepackage[utf8]{inputenc} 
\usepackage[T1]{fontenc}    
\usepackage{hyperref}       
\usepackage{url}            
\usepackage{booktabs}       
\usepackage{amsfonts}       
\usepackage{nicefrac}       
\usepackage{microtype}      
\usepackage{xcolor}         
\usepackage{adjustbox}
\usepackage{graphicx}
\usepackage{caption}
\usepackage{subcaption}
\usepackage{wrapfig}

\title{The Importance of Prompt Tuning for \\ Automated Neuron Explanations}

\author{Justin Lee\thanks{Shared First-authorship} \\
Mt. Carmel High School\\
\And
Tuomas Oikarinen\footnotemark[1] \\
UC San Diego
\And 
Arjun Chatha \\
Canyon Crest Academy
\And
Keng-Chi Chang\thanks{Equal contribution} \\
UC San Diego
\And
Yilan Chen\footnotemark[2] \\
UC San Diego 
\And
Tsui-Wei Weng \\
UC San Diego
}

\begin{document}

\maketitle

\begin{abstract}
  Recent advances have greatly increased the capabilities of large language models (LLMs), but our understanding of the models and their safety has not progressed as fast. In this paper we aim to understand LLMs deeper by studying their individual neurons. We build upon previous work showing large language models such as GPT-4 can be useful in explaining what each neuron in a language model does. Specifically, we analyze the effect of the prompt used to generate explanations and show that reformatting the explanation prompt in a more natural way can significantly improve neuron explanation quality and greatly reduce computational cost. We demonstrate the effects of our new prompts in three different ways, incorporating both automated and human evaluations.
\end{abstract}

\section{Introduction}
Large language models (LLMs) have exhibited remarkable capabilities across a variety of domains and tasks, such as text generation, question answering, and language translation. 
As an example, GPT-4 \cite{openai2023gpt4} exhibits human-level performance on various professional and academic benchmarks, including passing a simulated bar exam with a score around the top 10\% of test takers \cite{openai2023gpt4}. Even more, \cite{bubeck2023sparks} shows that GPT-4 can solve novel and difficult tasks that span mathematics, coding, vision, medicine, law, psychology, and more, viewing it as an early version of artificial general intelligence.

With the popularity of LLMs and their use in safety-critical and fairness-related applications such as healthcare \citep{singhal2023large, singhal2023towards}, education \citep{malinka2023educational, tan2023towards}, law \citep{openai2023gpt4, yu2022legal} and finance \citep{wu2023bloomberggpt, yang2023fingpt}, it is crucial to understand the models better and ensure their safety. 
Understanding how LLMs make their decisions can help us decide when to trust model predictions, detect bias in a network, and allow for greater control of the behavior the model exhibits. 
Recently, \cite{bills2023language} used GPT-4 to automatically write explanations for the behavior of neurons in large language models and to score those explanations, scaling an interpretability technique to all the neurons in a LLM. While impressive, their approach is still preliminary, and vast majority of the neurons cannot be explained well using this approach. This is in part due to some neurons simply not having a simple function that can be explained, but in part failures of the method to detect these roles.

In this paper, we improve on existing methods~\cite{bills2023language} to provide more accurate automated neuron explanations. Specifically, we propose 4 new and cost-effective prompts to improve the quality of single neuron explanations in LLMs. Our extensive experiments show that our proposed methods outperform the state-off-the-art~\cite{bills2023language} and can improve explanation performance over ~\cite{bills2023language} in terms of both automated and human evaluations, while being 2-3$\times$ more efficient and cost-effective.
%
%

\begin{figure}
    \centering
    \includegraphics[width=0.99\textwidth]{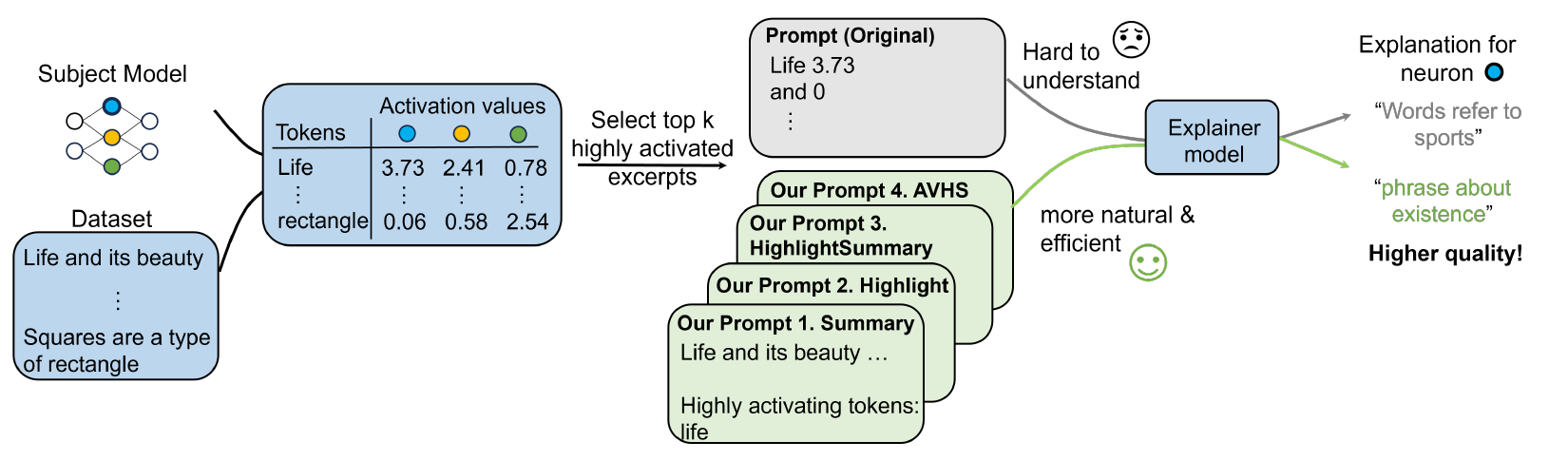}
    \caption{Overview of the neuron explanation pipeline and our proposed prompts (in green) to highly improve the explanation quality and efficiency.}
    \label{fig:overview}
\end{figure}

\section{Background and related work}

\textbf{Mechanistic and Neuron-level Interpretability.} 

A growing literature attempts to reverse engineer deep neural networks to understand the principles behind their operation \cite{olah2020zoom, elhage2021mathematical, olsson2022context}.
One foundation behind this approach is for humans to understand the function of individual neurons. However, given the sheer number of neurons in modern networks, having humans generate descriptions for each neuron is extremely labour intensive, even if the neurons are interpretable. 
To address this, some methods for generating automatic explanations have been proposed for vision models \cite{netdissect2017,hernandez2022natural,oikarinen2023clipdissect}.
Recently, a similar approach was proposed for language models, where \cite{bills2023language} use GPT-4 to automatically write explanations for the behavior of neurons in large language models (GPT-2) and score those explanations. In addition to \cite{bills2023language}, a host of previous method have been developed for understanding individual neurons in LLMs, but many of them are not automated or can only detect simpler concepts like single tokens. See \cite{sajjad2022neuronlevel} for a comprehensive overview of previous approaches.

\textbf{Explaining neurons in LLMs with GPT.}

In \cite{bills2023language}, the team from OpenAI showcases that GPT-4 can be useful in describing the roles of individual neurons in language models. Specifically, they focused on explaining the neurons in MLP layers of GPT2-XL\cite{radford2019language}, which is called the subject model (i.e. the model to be dissected and interpreted). The basic idea is to run a large text corpus $\mathcal{D}$ of text excerpts through the model, and record how highly each individual neuron activates for each token. In particular they used 60,000 random excerpts of 64 tokens each from the model's training data. They then find the 5 excerpts with the highest individual token activations for a specific neuron. These excerpts are then fed to the explainer model(GPT-4) together with the neuron's activation pattern following the prompt pattern shown in Fig.\ref{fig:prompt_overview}. The explainer model uses this information to generate a simple description of this neurons behavior. 

\textbf{GPT models.} GPT models are part of the broader family of transformer architectures \cite{vaswani2023attention}, which utilize attention mechanisms to process and generate text. 
A decoder-only transformer, such as GPTs, starts with a token embedding, followed by a series of ``residual blocks''. 
Each residual block contains an attention layer, followed by an MLP layer.
The attention layer computes weights for the model about which part of the input tokens to focus on and adjust the residuals streams accordingly.
The MLP layer then calculates activation based on the updated residual stream.
Finally, the residual stream is projected back to get the probability of next tokens. In this paper we focus on analyzing the neurons in the MLP layers.

\section{Methodology}

\begin{figure}
    \centering
    \includegraphics[width=\textwidth]{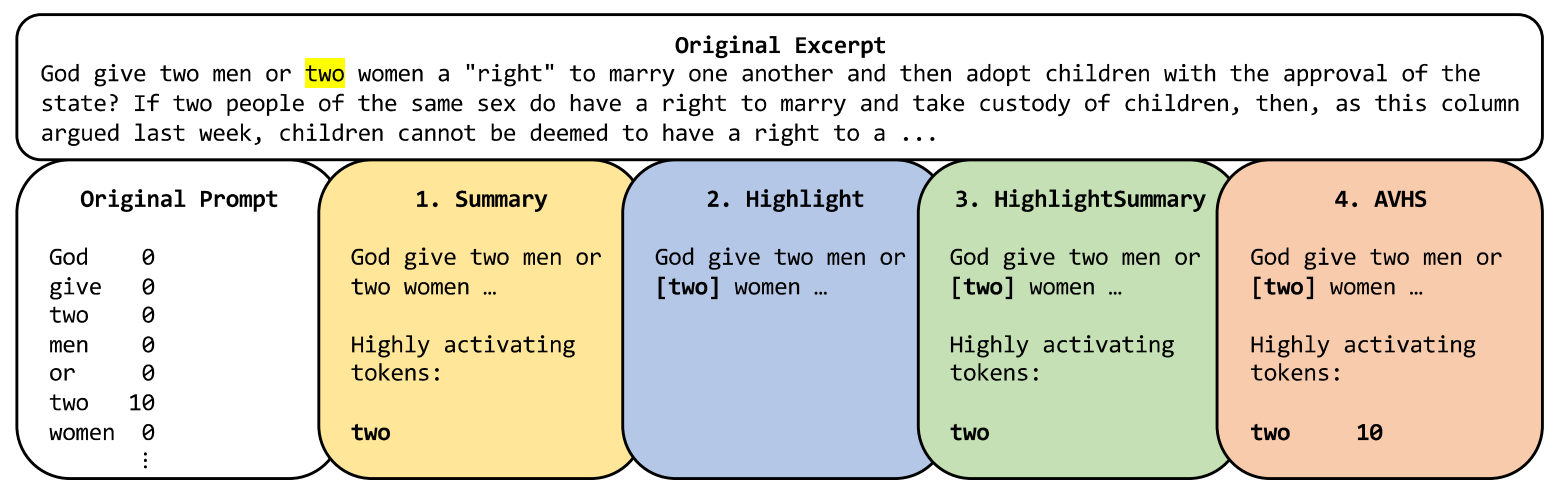}
    \caption{An overview of our proposed prompting methods, compared to the original prompt. Boldfacing is only done for the visual clarity, not part of actual prompt.} 
    \label{fig:prompt_overview}
\end{figure}

\textbf{Motivation.}
As introduced in the previous section, \cite{bills2023language} propose to use the \textit{explainer model} (GPT-4) to explain the neurons in the \textit{subject model} (GPT-2). To allow the explainer model to generate explanations of subject model neurons,  \cite{bills2023language} sends the explainer model a list of (token, activation) pairs separated by tabs and newlines as shows in Figure \ref{fig:prompt_overview}, where we call their method \textbf{Original prompt}, abbreviated as \textbf{Original}. 

However, we find that this kind of prompt has several drawbacks: (i) this requires 4 times as many tokens as the original text from $\mathcal{D}$, which results in large overhead and (ii) the text becomes unnatural/hard for a human to follow as it is interspersed with activation values all the time. 

To address these limitations, in this work we propose several simpler prompting methods, which have \textit{higher computation efficiency} and are more \textit{natural}. More importantly, we show in the experiments that our proposed prompts can improve the quality of neuron explanations compared to \cite{bills2023language}. 

\subsection{Proposed Approach}
Our approach builds on \cite{bills2023language} described in the previous section. Our goal is to test modifications to their pipeline to improve the quality and/or efficiency of the generated explanations. Specifically, we focused on changing the neuron explanation prompt given to explainer model.

Below we propose 4 new prompting strategies, which are showcased in Figure \ref{fig:prompt_overview} along with the \textbf{Original prompt} from \cite{bills2023language}:
\begin{enumerate}
    \item \textbf{Summary}: This prompt greatly simplifies the presentation, by just showing the original text excerpt and repeating a list of highly activating tokens (90\% quantile or above).
    \item \textbf{Highlight:} This is an alternative simplification, where we only show the original text but add square brackets around any highly activating tokens. 
    \item \textbf{HS (Highlight Summary)}: In this prompt, we combine our two approaches above, adding square brackets to highlight highly activating tokens in the text as well as a list of the highly activating tokens after the text excerpt. 
    \item \textbf{AVHS} (Activation Value + Highlight Summary): A combination of the previous prompts (with variations). This approach is similar to highlight summary, but also provides the values for the highly activating tokens -- which is similar but more compact than the original prompt as the original prompt also provides 0 activation values. Hence, this prompt provides the same amount of information as original but in a more concise and readable form.
\end{enumerate}

\subsection{Computational efficiency}

\label{sec:comp_efficiency}


\begin{wraptable}{r}{0.39\textwidth}
\vspace{-4mm}
\centering
\scalebox{0.88}{
\begin{tabular}{@{}lcc@{}}
\toprule
 & \begin{tabular}[c]{@{}l@{}}Tokens per\\  prompt\end{tabular} & Improvement \\ \midrule
Original~\cite{bills2023language} & 2338 & - \\
Summary & 959 & 2.44$\times$ \\
Highlight & \textbf{886} & \textbf{2.64$\times$} \\
HS & 1032 & 2.27$\times$ \\
AVHS & 1360 & 1.72$\times$ \\ \bottomrule
\end{tabular}
}
\caption{Average prompt length of different prompting methods. Our proposed prompts lead to 1.72-2.64$\times$ lower computatinal cost.}
\label{tab:comp_efficiency}
\vspace{-10mm}
\end{wraptable}

In Table \ref{tab:comp_efficiency}, we display the average prompt length needed for each neuron explanation(averaged over 50 neurons), including the few-shot examples given in the prompt.  As we can see, our prompts require 2-3$\times$ less tokens than original. This is important for a few reasons: first, the cost of generating explanations is calculated per token when using the API, so a 2$\times$ reduction in tokens per neuron mean you can evaluate 2$\times$ as many neurons for the same budget. 

Second, this could also lead to improvement in explanation performance. The prompt length is bounded by the model's context window, which is around 4096 tokens for most of the models we used. Since the length of original prompt is already close to this limit, it cannot be made much longer. With our prompts however, the prompt could be made longer by for example including additional few-shot examples, or giving the model more and/or longer text excerpts to summarize. This could improve the overall quality of the explanations.


\subsection{Evaluation methods}

\label{sec:simulate_and_score}

\textbf{Method 1: Simulate and Score.} The original OpenAI work \cite{bills2023language} mainly relied on simulation to test how good their explanations are. In this approach, after an explanation is generated a simulator model (also GPT-4 in their case) uses this explanation to predict a neuron's activations on unseen text excerpts. These predicted activations are then compared against actual neuron activations, and the explanation is scored based on the correlation between predicted activations and real activations. See \cite{bills2023language} for more details.
However, this method has a high computational cost, and we are unable to use GPT-4 as the simulator model as token probabilities are not available through the API. Nevertheless, we used simulation as one of our evaluation methods, utilizing the recently released \textit{GPT-3.5-Turbo-Instruct} model as the simulator, which provides good simulation accuracy at a reasonable cost.


\textbf{Method 2: Similarity to baseline explanation (AdaCS).}
As an alternative to the high computational cost of simulation, we developed a cheap and fast method so that we can compare different prompting methods at scale. To accomplish this, we decided to measure how similar our generated explanations are to provided explanations from OpenAI \cite{bills2023language}, available on \href{https://openaipublic.blob.core.windows.net/neuron-explainer/neuron-viewer/index.html}{NeuronViewer}. Since these explanations are not ground truth but generated by their method, we mostly restricted our attention to neurons with high simulation scores, which indicates their descriptions are highly accurate. The descriptions on NeuronViewer are generated using GPT-4 and include additional refining steps, so we expect them to be of higher quality than the descriptions we generate using GPT-3.5. Therefore, similarity to NeuronViewer description indicates the description is better. To compare the similarity of the descriptions, we use similarity in sentence embedding space. This is done using AdaV2 sentence embedding model from OpenAI, which we found to provide high quality similarity scores. Different prompting methods were then ranked based on how similar their descriptions are to the NeuronViewer explanation for that neuron, measured by cosine similarity of their embeddings. In addition to the real neurons, we used AdaCS to evaluate answers the different prompts provided to \textbf{Neuron Puzzles}. Neuron Puzzles are a set of artificial activation patterns created by \cite{bills2023language}, which were handcrafted to correspond to an interesting ground truth role for a neuron. In this case we are able to use AdaCS to measure the similarity of the generated description to the ground truth description. 

\textbf{Method 3: Human evaluation.} 
\label{sec:human_eval}
Finally we conducted a randomized study where the authors evaluated explanations for different neurons and compared explanations from different prompting methods. During the experiment, users had no way of knowing which explanation came from which method. The users were asked to evaluate 5 different explanations, and rate how good they were on a scale of 1-5, based on the neuron's activations visualized on NeuronViewer, as well as select which explanation was the best. An overview of our interface is shown in Figure \ref{fig:user_study_interface}.

\section{Experiments and Results}

\textbf{Setup.} In this section we evaluate the different prompting strategies using 4 different comparison methods which complement each other. As baseline, we used OpenAI's original code and prompt formatting to generate the explanations. We modified the explaining code available on \href{https://github.com/openai/automated-interpretability/tree/main}{GitHub} to experiment with our 4 new prompt formats in addition to the original. All prompts used the same few-shot examples as the original (but formatted according to the proposed new prompt).

\textbf{Explainer model:} We used both GPT-4 and GPT-3.5-turbo models as the explainer. While GPT-4 provides better explanations, it wasn't available at the start of this project and has a much higher usage cost. Thus we used GPT-3.5 for large scale experiments, and GPT-4 with smaller scale tests. See Table \ref{tab:total_cost} for an overview of our costs.

\textbf{Neurons evaluated:}
We followed \cite{bills2023language} and described neurons in the MLP layers of GPT-2(XL). This network contains 48 layers each with 6400 MLP neurons each, amounting to 307,200 neurons total. Due to limited computational budget, we evaluated several subsets of these neurons:

\begin{enumerate}
    \item \textbf{Random}: We randomly sampled an equal number of neurons from each of the 48 layers (20 per layer for GPT-3.5, 10 for GPT-4).
    \item \textbf{Random interpretable}: We randomly sampled neurons that had a score $>0.35$ on NeuronViewer, indicating these neurons are more interpretable (20 per layer for GPT-3.5, 10 for GPT-4). 
    \item \textbf{Top20 per layer}: We found the 20 neurons in each layer that had the highest simulation score, which is a proxy for most interpretable neurons of each layer.
    \item \textbf{Top 1k}: The 1,000 neurons in the model that had the highest simulation scores on NeuronViewer. These were mostly in the early layers as shown in Figure \ref{fig:top1k} in Appendix.
\end{enumerate}

In addition to randomly sampled neurons, we chose to focus on neurons that were somewhat well explained by \cite{bills2023language} as indicated by a high score on NeuronViewer. This was done to focus on more interesting and interpretable neurons.

\begin{table}[h!]
\centering
\scalebox{0.9}{
\begin{tabular}{@{}lccccccc@{}}
\toprule
 & \multicolumn{3}{c}{Explainer: GPT-3.5} &  & \multicolumn{3}{c}{Explainer: GPT-4} \\ 
\cline{2-4} \cline{6-8}
Prompt: & Random & \begin{tabular}[c]{@{}l@{}}Random Interp.\end{tabular} & Avg &  & Random & \begin{tabular}[c]{@{}l@{}}Random Interp.\end{tabular} & Avg \\
\midrule
Original \cite{bills2023language}  & 0.0962 & 0.2420 & 0.1718 &  & 0.1211 & \textbf{0.2284} & \textbf{0.1745} \\ 
Summary & \textbf{0.1249} & \textbf{0.2933} & \textbf{0.2123} &  & 0.1238 & 0.2252 & 0.1742 \\
Highlight & 0.1129 & 0.2603 & 0.1893 &  & \textbf{0.1241} & 0.2237 & 0.1737 \\
HS & 0.1239 & 0.2833 & 0.2065 &  & 0.1193 & 0.2173 & 0.1681 \\
AVHS & 0.1093 & 0.2791 & 0.1974 &  & 0.1219 & 0.2216 & 0.1715 \\ 
\bottomrule
\end{tabular}
}
\vspace{1em}
\caption{Simulate and score results. We can see summary performs clearly best for GPT-3.5, while GPT-4 results are quite similar across methods. Standard error of the mean was 0.0030-0.0041 for GPT-3.5 results and 0.0039-0.0055 for GPT-4 results. }
\label{tab:simulate}
\end{table}

\subsection{Simulate and score}

Table \ref{tab:simulate} displays the average scores across different settings when using simulate and score as described in section \ref{sec:simulate_and_score}. We can see Summary performs clearly the best when GPT-3.5 is the explainer model, with all our methods outperforming Original with statistical significance. When using GPT-4 on the other hand, Original performs the best on Random Interpretable neurons. However this is biased because Random Interpretable only selects neurons that originally received high simulation score (Using GPT-4 as the explainer and Original Prompt), and the effect goes away on complete random neurons. Finally, we note that the different methods perform more similarly when using GPT-4, with no statistically significant differences overall.

\subsection{AdaCS}

\begin{table}
\centering
\centering
\resizebox{0.96\textwidth}{!}{
\begin{tabular}{l|llllr}
\toprule
Prompt\textbackslash{}Neurons & Random & 
\begin{tabular}[c]{@{}l@{}}Random 
interpretable\end{tabular} & \begin{tabular}[c]{@{}l@{}}Top20 per layer\end{tabular} & Top 1k & \multicolumn{1}{l}{\textbf{Avg}} \\ 
\midrule
Original~\cite{bills2023language} & 0.8167 $\pm$ 0.0016 & 0.8369 $\pm$ 0.0020 & 0.8521 $\pm$  0.022 & 0.8820 $\pm$ 0.0025 & 0.8469 \\
Summary & 0.8471$\pm$ 0.0016 & \textbf{0.8790 $\pm$ 0.0015} & 0.8904 $\pm$ 0.0017 & \textbf{0.9026 $\pm$ 0.0019} & 0.8798 \\
Highlight & 0.8460 $\pm$ 0.0017 & 0.8700 $\pm$ 0.0017 & 0.8818 $\pm$ 0.0016 & 0.8930 $\pm$ 0.0021 & 0.8727 \\
HS & \textbf{0.8496 $\pm$ 0.0016} & 0.8782 $\pm$ 0.0015 & \textbf{0.8924 $\pm$ 0.0014} & 0.8995 $\pm$ 0.0020 & \textbf{0.8799} \\
AVHS & 0.8422 $\pm$ 0.0016 & 0.8667 $\pm$ 0.0018 & 0.8826 $\pm$ 0.0016 & 0.8961 $\pm$ 0.0015 & 0.8719\\
\bottomrule
\end{tabular}
}
\vspace{1em}
\caption{Average cosine similarity of generated explanations to baseline, using GPT-3.5 as the explainer model, as well as standard error of the mean. We can see all our methods noticeably outperform original.}
\label{tab:ada_cs}
\end{table}

We generated explanations for neurons from all the different settings above, explaining a total of 3880 neurons with GPT-3.5 as the explainer model. We then compared these to the GPT-4 produced explanations from \cite{bills2023language} using AdaCS, and report the results in Table \ref{tab:ada_cs}. We can see all our proposed prompts produce explanations significantly closer to GPT-4 explanations than original in all settings, with Summary and HighlightSummary performing the best.

Table \ref{tab:puzzles} shows the results on neuron puzzles, where the ground truth function is known, measured in AdaCS similarity of the produced explanation to ground truth. We can see our methods, especially Summary and HighlightSummary outperform Original quite clearly using both explainers. However due to only few puzzles being available, these results are on the edge of statistical significance.

\begin{wraptable}{r}{0.4\textwidth} 
\centering
\scalebox{0.9}{
\begin{tabular}{@{}lccc@{}}
\toprule
Prompt: & \multicolumn{1}{l}{\begin{tabular}[c]{@{}l@{}}Explainer: \\ GPT-3.5\end{tabular}}&  & \multicolumn{1}{l}{\begin{tabular}[c]{@{}l@{}}Explainer: \\ GPT-4\end{tabular}} \\ 
\midrule
Original \cite{bills2023language} & 0.8418 &  & 0.8560 \\ 
Summary & 0.8495 &  & \textbf{0.8657} \\
Highlight & 0.8414 &  & 0.8601 \\
HS & \textbf{0.8514} &  & 0.8636 \\
AVHS & 0.8490 &  & 0.8640\\
\bottomrule
\end{tabular}
}
\caption{Results on Neuron puzzles. Each score is average cosine similarity to ground truth explanation, across all 19 puzzles, with 3 explanations generated per puzzle per method. The standard error of the mean was 0.0042-0.0054 for all entries.}
\label{tab:puzzles}
\end{wraptable}

\subsection{Human evaluation}

We had 5 authors evaluate the descriptions provided by different prompts in a randomized setting as described in section \ref{sec:human_eval}. We had authors evaluate as judging description quality is often quite complex and requires close attention and expertise, making it not suitable for crowdsourcing. The results are shown in Table \ref{tab:user_study}. Each user rated 1 Randomly Interpretable neuron (>0.35 score) per layer, for a total of 240 evaluations per explainer model. We see that when using GPT-3.5 as the explainer, users largely preferred our new prompts, Summary being the best, followed HS, Highlight, AVHS and finally Original. When GPT-4 was used as the explainer instead, Summary was still rated the highest, but no explanation was better than others with statistical significance at this sample size. These findings indicate that simplifying the information provided to the model can lead to large gains, at least with less powerful explainer models, while the difference between prompts reduces when using GPT-4.

\begin{table}[h!]
\centering
\scalebox{0.9}{
\begin{tabular}{lccccc}
\toprule
 & \multicolumn{2}{c}{\begin{tabular}[c]{@{}c@{}}Explainer: GPT-3.5\end{tabular}} &  & \multicolumn{2}{c}{\begin{tabular}[c]{@{}c@{}}Explainer: GPT-4\end{tabular}} \\ 
\cline{2-3} \cline{5-6}
Prompt: & Avg rating: & \% chosen best &  & Avg rating: & \% chosen best \\ \midrule
Original~\cite{bills2023language} & 3.487 $\pm$ 0.075 & 14.17\% & & 4.367 $\pm$ 0.048 & 20.42\% \\
Summary & \textbf{4.308 $\pm$ 0.048} & \textbf{32.50\%} &  & \textbf{4.396 $\pm$ 0.047} & 20.42\% \\
Highlight & 4.054 $\pm$ 0.056 & 17.92\% & & 4.271 $\pm$ 0.054 & 17.92\% \\
HS & 4.196 $\pm$ 0.058 & 20.42\% &  & 4.350 $\pm$ 0.050 & \textbf{21.67\%} \\
AVHS & 3.842 $\pm$ 0.066 & 15.00\% &  & 4.312 $\pm$ 0.052 & 19.58\%\\
\bottomrule
\end{tabular}
}
\vspace{1em}
\caption{User study ratings averaged over evaluated neurons. We can see Summary performs the best for both explainer models, with large margin when using GPT-3.5. Avg rating shown together with standard error of the mean.}
\label{tab:user_study}
\end{table}

\section{Conclusions}

\begin{table}[b!]
\centering
\scalebox{0.9}{
\begin{tabular}{@{}lrrrrrrrr@{}}
\toprule
 & \multicolumn{1}{l}{\begin{tabular}[c]{@{}l@{}}GPT-3.5 \\ Simulate\end{tabular}} & \multicolumn{1}{l}{\begin{tabular}[c]{@{}l@{}}GPT-3.5\\ AdaCS\end{tabular}} & \multicolumn{1}{l}{\begin{tabular}[c]{@{}l@{}}GPT-3.5\\ Puzzles\end{tabular}} & \multicolumn{1}{l}{\begin{tabular}[c]{@{}l@{}}GPT-3.5\\ H. eval.\end{tabular}} & \multicolumn{1}{l}{\begin{tabular}[c]{@{}l@{}}GPT-4 \\ Simulate\end{tabular}} & \multicolumn{1}{l}{\begin{tabular}[c]{@{}l@{}}GPT-4\\ Puzzles\end{tabular}} & \multicolumn{1}{l}{\begin{tabular}[c]{@{}l@{}}GPT-4 \\ H. eval.\end{tabular}} & \multicolumn{1}{l}{Avg} \\ \midrule
Original & 5 & 5 & 4 & 5 & \textbf{1} & 5 & 2 & 3.86 \\
Summary & \textbf{1} & 2 & 2 & \textbf{1} & 2 & \textbf{1} & \textbf{1} & \textbf{1.43} \\
Highlight & 4 & 3 & 5 & 3 & 3 & 4 & 5 & 3.86 \\
HS & 2 & \textbf{1} & \textbf{1} & 2 & 5 & 3 & 3 & 2.43 \\
AVHS & 3 & 4 & 3 & 4 & 4 & 2 & 4 & 3.43\\
\bottomrule
\end{tabular}
}

\vspace{1em}
\caption{Summary of our results. Each value is the rank of the method according to that evaluation. We can see that across all evaluations, \textit{Summary} performs the best out of all prompting methods.}
\label{tab:summary}
\end{table}

We have summarized all our experimental results into Table \ref{tab:summary}. We can see that on average, the simple \textit{Summary} functions performs clearly the best, ranking first or second on every evaluation metric.  Based on these findings, and Table \ref{tab:comp_efficiency} showing \textit{Summary} is the second most computationally efficient and $2.44\times$ more computationally efficient than original, we think it is the best choice for neuron explanations. Interestingly additional information from HS or AVHS was somewhat harmful to model performance, suggesting that explainer models struggle to handle too much complexity.




\section*{Acknowledgement}
The authors would like to thank San Diego Supercomputer Center, Halicioglu Data Science Institute and NSF's ACCESS program for computing support. T. Oikarinen and Y. Chen are supported by National Science Foundation under Grant No. 2107189. T.-W. Weng is supported by National Science Foundation under Grant No. 2107189 and 2313105.

\newpage
\clearpage

\bibliographystyle{plain}
\bibliography{main}




\clearpage
\newpage

\appendix

\section{Appendix}

\subsection{Total cost and computation}

Table \ref{tab:total_cost} displays an estimated computational cost and time for different steps of our research, when using the OpenAI API. These costs are for running all 5 prompting methods.

\begin{table}[h]
\centering
\begin{tabular}{llll} 
\toprule
Model& Setting& Cost per 1k neurons & Runtime per 1k neurons\\ \hline
GPT-3.5-Turbo& Explain& \$2.50& 3 hr\\
GPT-3.5-Turbo-Instruct& Simulate+Score& \$80& 6 hr\\
GPT-4& Explain& \$200& 12 hr\\
Ada& Compare& \$0.02& 1 hr\\
\bottomrule
\end{tabular}
\vspace{1em}
\caption{Cost and runtime estimates in different settings}
\label{tab:total_cost}
\end{table}

\subsection{Additional figures}

\begin{figure}[h]
     \centering
     \begin{subfigure}[b]{0.49\textwidth}
         \centering
         \includegraphics[width=\textwidth]{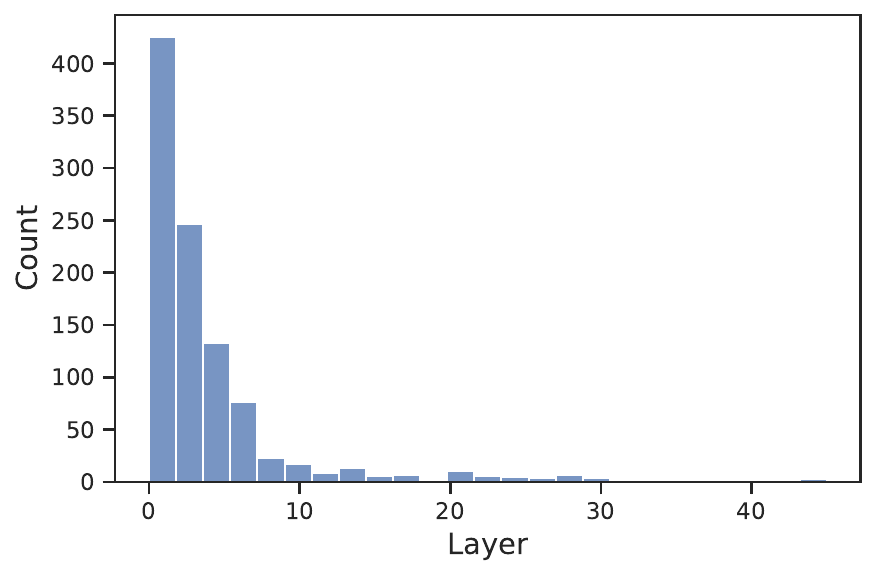}
         \caption{Distribution of top 1k neurons by layers}
         \label{fig:neuron_distribution}
     \end{subfigure}
     \begin{subfigure}[b]{0.49\textwidth}
         \centering
         \includegraphics[width=\textwidth]{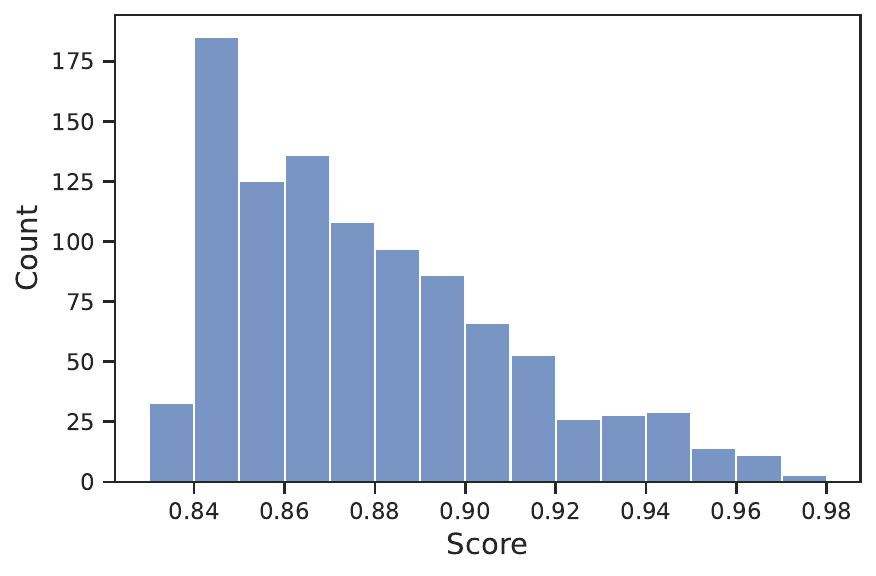}
         \caption{Distribution of top 1k neurons by scores}
         \label{fig:neuron_score_top1k}
     \end{subfigure}
    \caption{Distribution of the Top 1k neurons by layers and scores}
    \label{fig:top1k}
\end{figure}

Figure \ref{fig:neuron_distribution} shows the distribution of topk1k neurons, i.e. the ones with highest explanations scores according to \cite{bills2023language}. The majority of Top 1k neurons comes from the first few layers of GPT according. There is a notable spike in Layer 1 (layers are start from 0) meaning that the activating texts in Layer 1 are on average easier to explain. The scores for top1k range from 0.83 to 0.98. 

\begin{figure}
    \centering
    \includegraphics[width=0.95\textwidth]{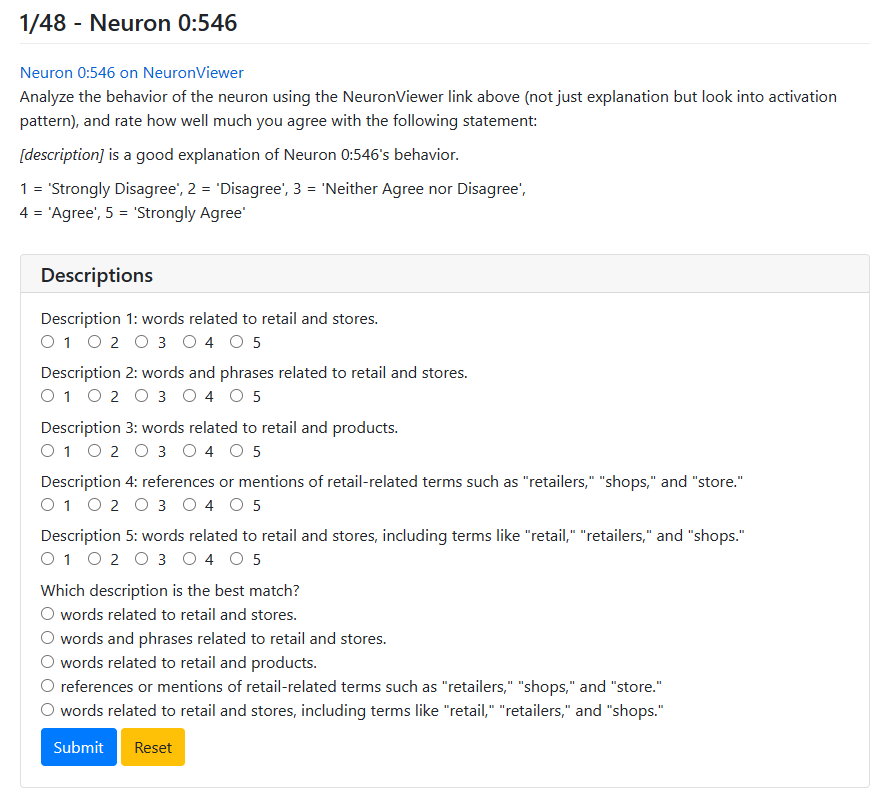}
    \caption{User study interface. To be viewed together with Neuron's activation patterns on NeuronViewer.}
    \label{fig:user_study_interface}
\end{figure}

Figure \ref{fig:user_study_interface} shows the interface used for our user study. 

\end{document}